\documentclass[conference, 10pt]{IEEEtran}

\usepackage{cite}
\usepackage{url}
\usepackage{graphicx}
\usepackage{color}
\usepackage{placeins}
\usepackage{float}
\usepackage{tabularx,colortbl}
\usepackage{ifthen}
\usepackage{booktabs} 
\usepackage{subcaption}

\makeatletter

\newcounter{author}
\renewcommand{\author}[2][]{
   \stepcounter{author}
   \@namedef{author@\theauthor}{#2}
   \@namedef{authorlabel@\theauthor}{#1}
}

\newcounter{address}
\newcommand{\address}[2][]{
   \stepcounter{address}
   \@namedef{address@\theaddress}{#2}
   \@namedef{addresslabel@\theaddress}{#1}
}

\newcommand{\alsep}{and}

\def\newmaketitle{\par%
  \begingroup%
  \normalfont%
  \def\thefootnote{}
  \def\footnotemark{}
  \let\@makefnmark\relax
  \footnotesize
  \footnotesep 0.7\baselineskip
  \normalsize%
  \twocolumn[\thenewmaketitle\@IEEEaftertitletext]%
  \if@IEEEusingpubid
     \enlargethispage{-\@IEEEpubidpullup}%
  \fi
  \endgroup
  \setcounter{footnote}{0}\let\maketitle\relax\let\@maketitle\relax
  \gdef\@thanks{}%
  \let\thanks\relax}

\def\thenewmaketitle{
  \newpage
  \begin{center}%
    \vskip0.2em{\Huge\@IEEEcompsoconly{\sffamily}\@IEEEcompsocconfonly{\normalfont\normalsize\vskip 2\@IEEEnormalsizeunitybaselineskip
    \bfseries\large}\@title\par}\vskip1.0em\par%
     \vspace{1ex}
     \newcounter{c@author}
     \newcounter{c@tmp}
     \ifthenelse{\value{author}=2}{%
       \newcommand{\liand}{ and }}{%
       \newcommand{\liand}{, and }}
     \ifthenelse{\value{address}<2}{%
       \@nameuse{author@1}%
       \stepcounter{c@author}%
       \whiledo{\value{c@author}<\value{author}}{%
         \setcounter{c@tmp}{\value{author}}%
         \addtocounter{c@tmp}{-\value{c@author}}%
         \ifthenelse{\value{c@tmp}=1}{%
           \renewcommand{\alsep}{\liand}}{\renewcommand{\alsep}{, }}%
         \stepcounter{c@author}\alsep \@nameuse{author@\thec@author}}\\%
     }
     {
       \@nameuse{author@1}${}^{(\ref{\@nameuse{authorlabel@1}})}$%
       \stepcounter{c@author}%
       \whiledo{\value{c@author}<\value{author}}{%
       \setcounter{c@tmp}{\value{author}}%
       \addtocounter{c@tmp}{-\value{c@author}}%
       \ifthenelse{\value{c@tmp}=1}{%
         \renewcommand{\alsep}{\liand}}{\renewcommand{\alsep}{, }}%
       \stepcounter{c@author}\alsep \@nameuse{author@\thec@author}%
         ${}^{(\ref{\@nameuse{authorlabel@\thec@author}})}$%
       }
     }
     \vspace{0.2ex}
     \ifthenelse{\value{address}>0}{%
       \ifthenelse{\value{address}=1}{
         {\@nameuse{address@1}}
       }
       {
         \newcounter{c@address}
         \begin{center}
         \whiledo{\value{c@address}<\value{address}}
         {
           \refstepcounter{c@address}
             ${}^{(\thec@address)}$\,%
               \label{\@nameuse{addresslabel@\thec@address}}%
               \@nameuse{address@\thec@address}\\ %
         }
         \end{center}
       } 
     }
     {
       \relax
     }
  \end{center}
}

\makeatother

\title{Hemispherical Angular Power Mapping of Installed mmWave Radar Modules Under Realistic Deployment Constraints}

\author[org1]{Maaz Qureshi}
\author[org1]{Mohammad Omid Bagheri}
\author[org1]{William Melek}
\author[org1]{George Shaker}

\address[org1]{The University of Waterloo, Waterloo, Ontario N2L 3G1, Canada, http://uwaterloo.ca}

\begin{document}

\newmaketitle

\begin{abstract}
Characterizing the angular radiation behavior of installed millimeter-wave (mmWave) radar modules is increasingly important in practical sensing platforms, where packaging, mounting hardware, and nearby structures can significantly alter the effective emission profile. However, once a device is embedded in its host environment, conventional chamber- and turntable-based antenna measurements are often impractical. This paper presents a hemispherical angular received-power mapping methodology for in-situ EM validation of installed mmWave modules under realistic deployment constraints. The approach samples the accessible half-space around a stationary device-under-test by placing a calibrated receiving probe at prescribed $(\phi,\theta,r)$ locations using geometry-consistent positioning and quasi-static acquisition. Amplitude-only received-power is recorded using standard RF instrumentation to generate hemispherical angular power maps that capture installation-dependent radiation characteristics. Proof-of-concept measurements on a 60-GHz radar module demonstrate repeatable hemispherical mapping with angular trends in good agreement with full-wave simulation, supporting practical on-site characterization of embedded mmWave transmitters.
\end{abstract}

\section{Introduction}

Millimeter-wave (mmWave) radar modules are rapidly emerging as key sensing technologies in vehicles, aerial platforms, industrial automation, and biomedical/wearable systems \cite{bagheri2024radar,hasch2012millimeter,bagheri2024metasurface}. In these applications, the module typically operates embedded within a host platform rather than as an isolated radiator. Consequently, installation effects, such as dielectric loading from covers and radomes, nearby metallic features, and platform scattering, can distort the angular received-power distribution and shift directionality from free-space expectations \cite{abedi2024use}. This motivates practical validation approaches that characterize mmWave transmitters in their installed configuration under realistic deployment conditions.

Conventional radiation characterization typically assumes access to dedicated measurement infrastructure, where the device-under-test (DUT) is aligned and rotated in a controlled range using anechoic chambers, turntables, and far-field procedures \cite{fordham2022ieee,bagheri2025dielectric}. While these techniques provide high fidelity for standalone antennas, they are often difficult to apply to embedded mmWave radar modules integrated into complex platforms. Moreover, removing an installed module for laboratory testing can alter its electromagnetic bound ary conditions, while repeated re-mounting and alignment introduce uncertainty that obscures installation-dependent behavior. These limitations motivate alternative methodologies that enable repeatable in-situ angular characterization without requiring platform rotation or specialized facilities, particularly in constrained or semi-controlled environments \cite{froes2019antenna,11293594}.

Hemispherical sampling provides a practical geometry for angular characterization of embedded mmWave radar modules because it aligns with the accessible observation region of
most installed platforms \cite{exposito2023car}. Consequently, full-sphere pattern capture is often unnecessary or physically infeasible, since the obscured region is either inaccessible or outside the operational field-of-view of the sensing system. A hemispherical measurement surface centered on the module and oriented toward the accessible region therefore provides a physically meaningful and application-relevant basis for angular evaluation. This geometry enables consistent angular indexing while keeping the DUT fixed and allows received-power mapping over the application-relevant observation region without platform rotation. The resulting hemispherical maps reveal installation-induced angular energy redistribution due to loading, masking, and scattering.

This paper presents a hemispherical received-power mapping methodology for in-situ EM characterization of installed mmWave radar modules under realistic deployment constraints. By sampling the accessible half-space using amplitude-only measurements and standard RF instrumentation, the proposed approach provides a practical means to evaluate installation-dependent angular behavior without relying on turntable-based facilities.

The main contributions of this work are summarized as follows: (i) In-situ hemispherical power mapping: A measurement methodology to characterize the angular received-power distribution of platform-integrated mmWave modules while preserving the installed configuration. (ii) Deployment-aware measurement workflow: A geometry-consistent sampling framework that enables hemispherical scans around mechanically constrained devices using automated probe positioning as a measurement tool, without relying on turntables or full-range infrastructure. (iii) Validation at 60~GHz: Experimental results demonstrating repeatable angular power maps and good agreement with full-wave simulation trends, supporting practical EM evaluation under realistic deployment constraints.

\section{Measurement Concept and Deployment Requirements}

The objective of this work is to enable practical in-situ angular characterization of installed mmWave radar modules when conventional antenna test ranges are infeasible. Rather than targeting metrology-grade far-field patterns, the proposed approach performs hemispherical received-power mapping around a stationary DUT while preserving installed boundary conditions. The output is an angularly indexed power map that captures installation-dependent directionality influenced by nearby structures, packaging materials, and platform scattering.

\begin{figure}[!t] 
\centering
\includegraphics[width=3.4in]{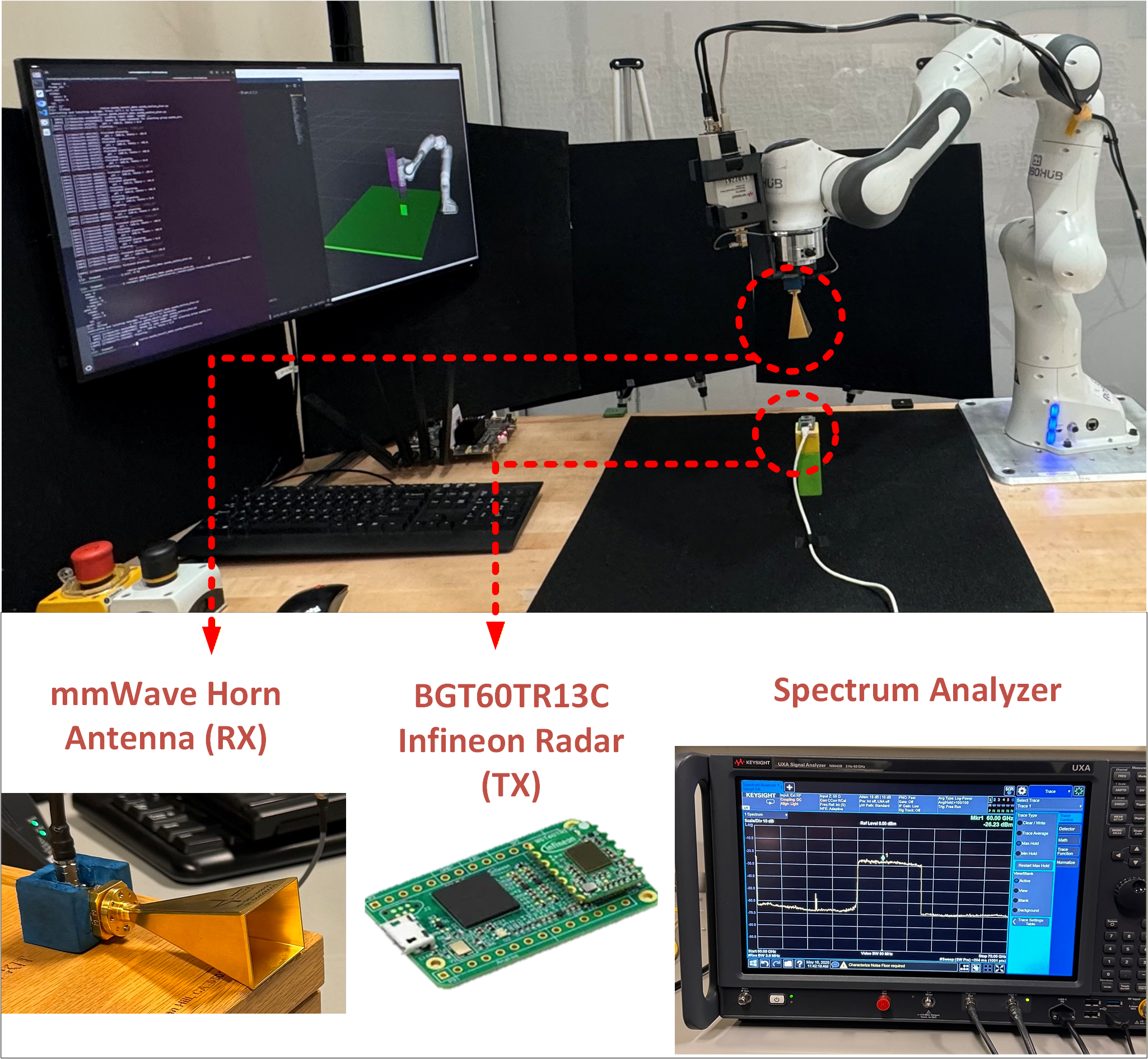}
\caption{Semi-anechoic hemispherical mapping setup for in-situ characterization of an installed mmWave radar module, including automated probe positioning, stationary DUT, absorber-lined workspace, and spectrum-analyzer-based received-power logging for angular power map generation.}
\label{fig:system_setup}
\vspace{-1mm}
\end{figure}

The core concept is to sample a hemispherical surface around the DUT and record amplitude-only received-power at prescribed observation angles. A calibrated mmWave horn probe is positioned at $(\phi,\theta,r)$ locations, where $\phi$ and $\theta$ denote the observation direction and $r$ is the measurement radius. The resulting samples form a hemispherical power map of the installed module. This power-domain mapping preserves DUT boundary conditions and enables repeatable angular characterization without platform rotation.

The proposed methodology is designed to meet practical deployment constraints that arise in industry-oriented validation of embedded mmWave modules. In realistic applications, the DUT is often integrated into mechanically constrained assemblies such as automotive fascia regions, robotic housings, structural frames, or protective enclosures. These environments introduce limitations that directly shape the measurement strategy: (i) No DUT rotation: The host platform may be too large, heavy, or mechanically fixed to rotate on a turntable; thus, the measurement must be performed with the DUT stationary. (ii) Limited accessible field-of-view: Surrounding structures may block a significant portion of the full sphere, motivating hemispherical sampling over the application-relevant half-space. (iii) Constrained scanning volume: Probe motion must respect clearance constraints near the platform while maintaining a consistent measurement radius. (iv) Non-dedicated environment: Measurements may be conducted in a semi-anechoic lab or engineering workspace; the method should remain robust to residual multipath and partial absorber coverage. (v) Portability and setup simplicity: The workflow should require minimal infrastructure, using standard instrumentation and repeatable geometry calibration. (vi) Repeatability: Consistent angular indexing is required across repeated scans to support configuration comparisons.

To satisfy these requirements, the system adopts an automated probe-positioning workflow that enables hemispherical sampling around a stationary DUT, combined with amplitude-only received-power acquisition. This design emphasizes accessibility and repeatability for installed-module evaluation, while producing angular power maps that directly support engineering decisions such as field-of-view verification, packaging impact assessment, and installation troubleshooting.


\section{Experimental Setup and Hemispherical Mapping Results}
The proposed measurement platform, shown in Fig.~\ref{fig:system_setup}, enables repeatable hemispherical received-power mapping around a stationary DUT under realistic deployment constraints. The system combines automated probe positioning, a geometry-calibrated end-effector, and synchronized RF acquisition to support consistent angular sampling without rotating the DUT.

\subsection{Measurement Setup}
A proof-of-concept hemispherical received-power mapping experiment was conducted using a commercial 60-GHz mmWave radar module as the DUT. The module was fixed in a representative installed configuration to preserve platform boundary conditions. A calibrated receiving antenna was positioned at a constant radius around the DUT to sample the accessible half-space while maintaining safe clearance from nearby structures.

Automated probe positioning is implemented using a 7-DoF Franka Emika Panda manipulator to support repeatable hemispherical sampling. The end-effector mounts a WR-15 horn probe and harmonic mixer for direct downconversion at the probe tip, reducing mmWave cable loss, while received-power magnitude is recorded by a spectrum analyzer. Key system parameters are summarized in Table~\ref{tab:system_params}.

\begin{figure*}[t!]
\centering

\begin{subfigure}[t]{0.32\linewidth}
    \centering
    \includegraphics[width=\linewidth]{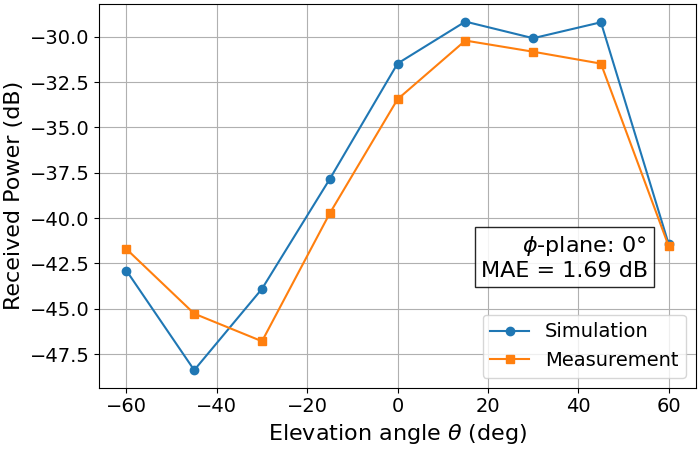}
    \caption{}
\end{subfigure}
\begin{subfigure}[t]{0.32\linewidth}
    \centering
    \includegraphics[width=\linewidth]{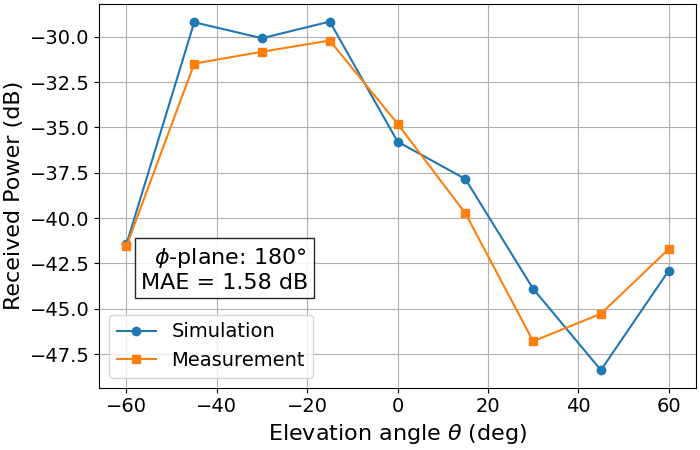}
    \caption{}
\end{subfigure}
\begin{subfigure}[t]{0.32\linewidth}
    \centering
    \includegraphics[width=\linewidth]{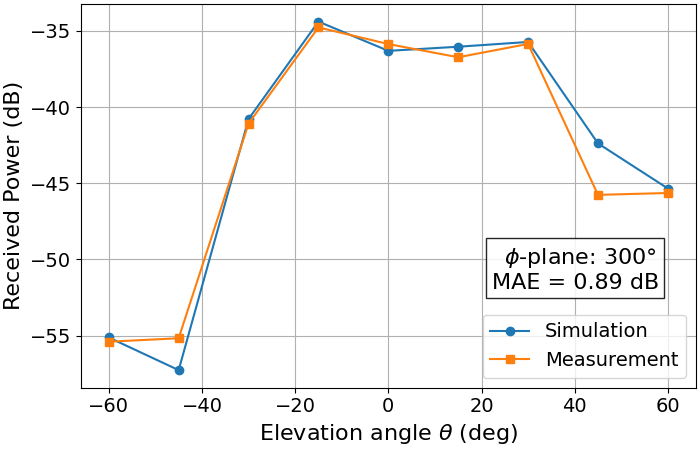}
    \caption{}
\end{subfigure}

\vspace{2mm}

\begin{subfigure}[t]{0.32\linewidth}
    \centering
    \includegraphics[width=\linewidth]{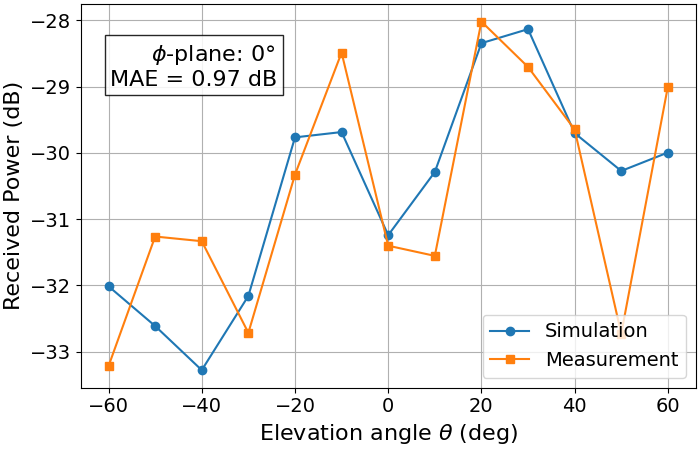}
    \caption{}
\end{subfigure}
\begin{subfigure}[t]{0.32\linewidth}
    \centering
    \includegraphics[width=\linewidth]{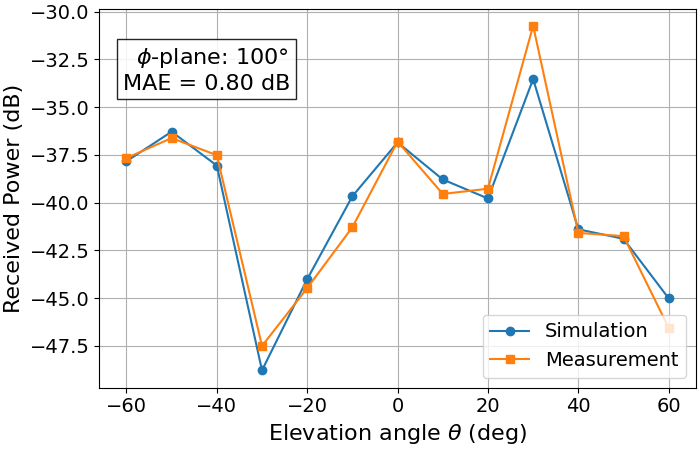}
    \caption{}
\end{subfigure}
\begin{subfigure}[t]{0.32\linewidth}
    \centering
    \includegraphics[width=\linewidth]{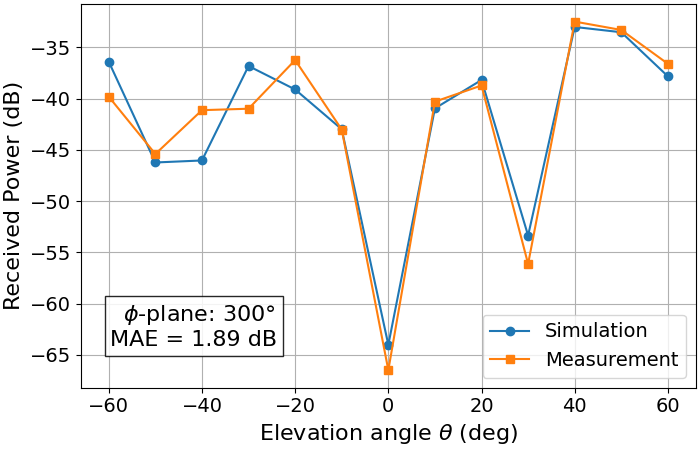}
    \caption{}
\end{subfigure}

\caption{Comparison of normalized received-power in simulation and measurement at the workspace boundary. (a)–(c) correspond to $r=8$~cm with $15^\circ$ sampling for $\phi$-plane cuts at $0^\circ$, $180^\circ$, and $300^\circ$, and (d)–(f) correspond to $r=5$~cm with $10^\circ$ sampling for $\phi$-plane cuts at $0^\circ$, $100^\circ$, and $300^\circ$.}
\label{fig:polar_results}
\end{figure*}

Unlike manual measurements that rely on operator-dependent alignment, the proposed workflow employs a geometry-calibrated transformation chain. The sampling grid is defined in spherical coordinates $(\phi,\theta,r)$ with respect to the DUT center and mapped into the positioning system base frame through a kinematic transform $T_{\mathrm{final}}$, ensuring consistent probe orientation toward the DUT across all sampling points:
\begin{equation}
T_{\mathrm{final}} = T_{\mathrm{base}}\times T(\phi,\theta,r)\times T_{\mathrm{offset}}
\end{equation}
where $T_{\mathrm{base}}$ defines the robot base frame, $T(\phi,\theta,r)$ maps the spherical sampling grid to the desired probe pose relative to the DUT, and $T_{\mathrm{offset}}$ applies the calibrated end-effector mounting offset.

The scan trajectory is generated using a collision-aware motion planner that accounts for the DUT platform and surrounding workspace geometry. To ensure stable mmWave acquisition, measurements are performed in a quasi-static manner: the probe moves to the target pose, dwells for a short stabilization period (1--2~s), and triggers RF acquisition after residual vibrations have decayed.

\begin{table}[t!]
\caption{Measurement System Parameters}
\label{tab:system_params}
\begin{center}
\renewcommand{\arraystretch}{1.2} 
\begin{tabular}{lc}
\toprule
\textbf{Parameter} & \textbf{Value} \\
\midrule
Operating Frequency & 58 -- 63 GHz \\
Scan Radius ($r$) & 3 cm -- 15 cm \\
Polar Range ($\theta$) & $0^{\circ}$ -- $70^{\circ}$ \\
Angular Step ($\Delta\phi, \Delta\theta$) & $10^{\circ}$ (Fine) -- $20^{\circ}$ (Coarse) \\
Dwell Time per Point & 1 -- 2 s \\
Robot Repeatability & $\pm 0.1$ mm (Nominal) \\
\bottomrule
\end{tabular}
\end{center}
\vspace{-2mm}
\end{table}

Amplitude-only received-power was recorded at each sampling direction using standard RF instrumentation. Measurements were performed in a semi-anechoic workspace with localized absorbers to suppress dominant reflections, enabling a deployable workflow without requiring full anechoic ranges or turntable-based facilities. To enable in-situ testing without a full anechoic chamber, the measurement workspace is augmented with RF absorber sheets placed on the table surface and overhead. This localized absorber treatment mitigates dominant specular reflections and creates a compact quiet zone around the DUT suitable for power-domain characterization.


\begin{table}[t!]
\caption{Summary of Measurement Performance}
\label{tab:performance_summary}
\begin{center}
\renewcommand{\arraystretch}{1.2} 
\begin{tabular}{lcc}
\toprule
\textbf{Metric} & \textbf{Manual Baseline} & \textbf{Robotic System} \\
\midrule
Mean Absolute Error & $\approx 1.8$ -- $2.2$ dB & $\approx \textbf{1.2}$ -- $\textbf{1.6}$ \textbf{dB} \\
Intra-Day Repeatability & N/A (High Variability) & $\mathbf{< 0.20}$ \textbf{dB} \\
Scan Coverage Success & Operator Dependent & \textbf{100\%} \\
\bottomrule
\end{tabular}
\end{center}
\vspace{-2mm}
\end{table}


\subsection{Hemispherical Angular Power Mapping Results}
To demonstrate feasibility under realistic constraints, A proof-of-concept experiment was conducted using an Infineon BGT60TR13C (58–63 GHz) mmWave radar module as the DUT mounted in a fixed bench-top installation. Hemispherical scans were conducted around the stationary DUT at multiple near-field radii ($r=4$--15~cm). The angular step size was adapted between $10^{\circ}$ and $20^{\circ}$ to maintain sufficient spatial sampling density for capturing both main-lobe and side-lobe features.

The measured data were compiled into hemispherical angular power maps indexed by $(\phi,\theta)$, providing a compact representation of the installed module’s directional behavior over the accessible field-of-view. Across repeated scans, the workflow produced consistent angular trends, enabling comparative installation studies (e.g., covers, nearby structures, or mounting locations). The automated measurements were benchmarked against an idealized full-wave simulation of the radar package and compared with a manual probe-positioning baseline. As shown in Table~\ref{tab:performance_summary}, the proposed workflow improves measurement fidelity by reducing operator-induced variability and achieving lower mean absolute error (MAE) relative to the simulation reference.

To provide a physics-based reference, a full-wave electromagnetic simulation of the DUT configuration was performed. The measured hemispherical maps showed good qualitative agreement with simulated angular trends, capturing the dominant regions of higher and lower received-power. This agreement supports hemispherical received-power mapping as a practical validation tool for installed modules when the objective is to assess deployment-dependent directionality rather than perform metrology-grade far-field characterization.

Figure~\ref{fig:polar_results} demonstrates the angular consistency of the proposed hemispherical mapping by comparing simulated and measured received-power across multiple $\phi$-plane cuts at two near-field radii. Subplots (a)–(c) correspond to $r=8$~cm with $15^\circ$ sampling, where the measured curves closely follow the simulated profiles and preserve the dominant rise-and-fall behavior across elevation, yielding MAE values on the order of $\sim$1--2~dB. Subplots (d)–(f) show the more challenging near-field case at $r=5$~cm with finer $10^\circ$ sampling, where the measurements still track the main angular trends but exhibit increased local fluctuations and sharper variations in some cuts due to stronger sensitivity to small geometric offsets and installation-dependent interactions at shorter range.

\subsection{Practical Implications Under Deployment Constraints}
The results indicate that hemispherical received-power mapping provides a practical middle ground between manual probing and facility-based antenna measurements for installed mmWave transmitters. By sampling the accessible half-space around a stationary DUT, the workflow preserves installation-dependent effects (loading, masking, and scattering) while enabling geometry-consistent scans using standard instrumentation in semi-controlled environments without dedicated anechoic facilities.

Collision-free trajectory planning enabled dense hemispherical sampling across all tested radii without contacting the DUT or surrounding structures, supporting repeatable scans in clearance-constrained installations.

The measured maps reflect the characteristics of the full measurement chain, including probe response and residual multipath in non-dedicated environments. However, the objective is not to replace classical antenna metrology; rather, the proposed approach provides an accessible and repeatable method for identifying installation-driven angular behavior and comparing platform configurations using a consistent hemispherical sampling geometry. This makes the methodology well suited for industry-oriented validation tasks such as field-of-view verification, packaging impact assessment, and installation troubleshooting of embedded mmWave radar modules.

\section{Conclusion}
This paper presented a hemispherical angular received-power mapping methodology for in-situ EM validation of installed mmWave radar modules under realistic deployment constraints. The approach samples the accessible half-space around a stationary DUT and records amplitude-only received-power using standard RF instrumentation, producing angular power maps that capture installation-dependent behavior without requiring DUT rotation or full-range anechoic facilities. Proof-of-concept measurements at 60~GHz demonstrated repeatable mapping with angular trends in good agreement with full-wave simulation, supporting on-site characterization of embedded mmWave transmitters. Future work will focus on accelerating scan time, expanding evaluation across richer installation scenarios and platform geometries, and extending the framework toward coherent acquisition and advanced post-processing for deeper diagnostic insight. Overall, the workflow is well suited for engineering validation tasks such as field-of-view verification, packaging impact assessment, and installation troubleshooting of embedded mmWave modules.


\section{Acknowledgments} \label{sec:VIII}
The authors acknowledge Keysight Technologies for their support. This work was also supported, in part, by Infineon, Google, Rogers, and MITACS.

\bibliographystyle{IEEEtran}
\bibliography{Ref}

\end{document}